\documentclass[letterpaper]{article} %
\usepackage[preprint]{aaai2027}      %
\usepackage[hyphens]{url}    %
\usepackage{graphicx}        %
\usepackage{natbib}          %
\usepackage{caption}         %
\usepackage{algorithm}
\usepackage{algorithmic}
\usepackage{newfloat}
\usepackage{listings}
\DeclareCaptionStyle{ruled}{labelfont=normalfont,labelsep=colon,strut=off}
\floatstyle{ruled}
\newfloat{listing}{tb}{lst}{}
\floatname{listing}{Listing}
\usepackage{makecell}

\usepackage{amsmath}         %
\usepackage{amssymb}         %
\usepackage{booktabs}        %
\usepackage{multirow}
\usepackage{xcolor}
\usepackage{tikz}
\usetikzlibrary{arrows.meta,positioning,shapes.geometric,calc,fit,backgrounds,decorations.pathreplacing}
\usepackage{fontawesome5}

\definecolor{cClinical}{HTML}{2563EB}
\definecolor{cReports}{HTML}{B45309}
\definecolor{cRNA}{HTML}{059669}
\definecolor{cMut}{HTML}{DC2626}
\definecolor{cCNV}{HTML}{7C3AED}
\definecolor{cHisto}{HTML}{4338CA}
\definecolor{cPool}{HTML}{0F766E}
\definecolor{cTeacher}{HTML}{7C3AED}
\definecolor{cStudent}{HTML}{0D9488}
\definecolor{cDoctor}{HTML}{059669}
\definecolor{cBrain}{HTML}{DB2777}
\definecolor{cGold}{HTML}{D97706}
\definecolor{cKey}{HTML}{EAB308}
\definecolor{cPatient}{HTML}{2563EB}

\newcommand{\bx}{\mathbf{x}}

\newcommand{\bmm}{\mathbf{m}}

\newcommand{\E}{\mathbb{E}}
\newcommand{\R}{\mathbb{R}}

\newcommand{\Softmax}{\mathrm{softmax}}
\newcommand{\Argmax}{\operatorname*{argmax}}
\newcommand{\Argmin}{\operatorname*{argmin}}

\newcommand{\ourmethod}{\textsc{OracleImitator}}

\newcommand{\fitcol}[1]{\resizebox{\ifdim\width>\columnwidth\columnwidth\else\width\fi}{!}{#1}}
\newcommand{\fittext}[1]{\resizebox{\ifdim\width>\textwidth\textwidth\else\width\fi}{!}{#1}}

\title{Active Feature Acquisition by Distilling an Oracle Teacher}

\author{
  Yang Meng,
  Zixin Ding,
  Robert L. Grossman,
  Yuxin Chen
}
\affiliations{
  University of Chicago
}

\begin{document}
\maketitle

\begin{abstract}
Acquiring every feature of an instance is often costly: a measurement can
carry a monetary price, a time delay, or a privacy cost. Active Feature
Acquisition (AFA) addresses this by learning a policy that decides, per
instance, which features to acquire under a budget. Existing AFA methods train
this policy by optimizing a self-constructed utility target rather than
distilling an explicit teacher. We construct this teacher explicitly. At
training time the label and the unobserved feature values are available, and
together they determine the teacher in closed form: an oracle that ranks every
candidate by its effect on the downstream predictor's error. \ourmethod\
distills this ranking into a deployable policy by soft-target imitation. Across
multimodal medical diagnosis and the AFABench benchmark, \ourmethod\ matches or
outperforms existing AFA methods across the budget range, recovering $90\%$ of the attainable accuracy gain at under a quarter of the acquisition cost. Because the teacher is an explicit algorithmic object rather than an implicit optimization target, we can also
measure directly how much of its ranking is recoverable by a policy restricted
to what is visible at deployment.
\end{abstract}

\section{Introduction}
\label{sec:intro}

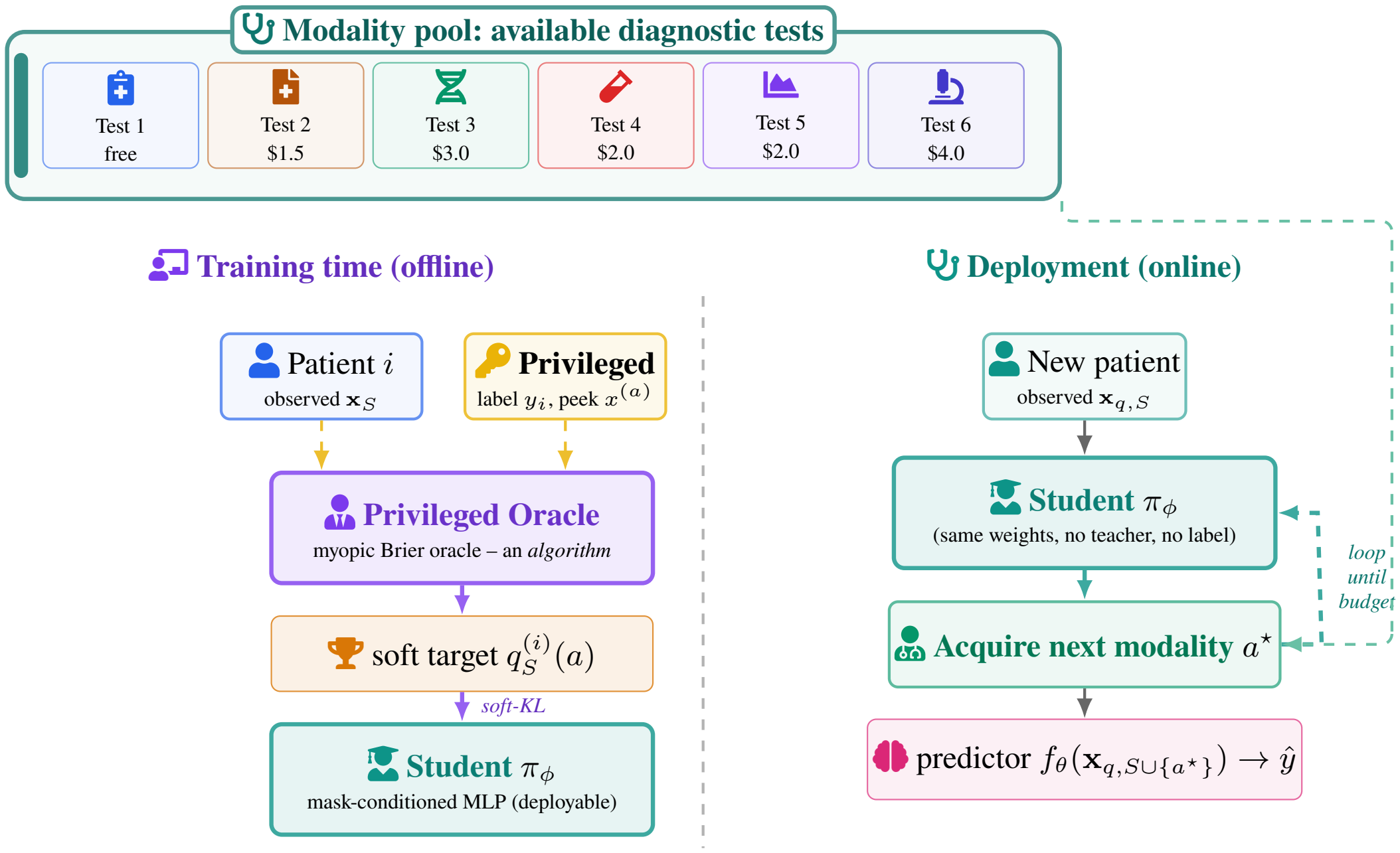
\begin{figure*}[t]
\centering
\definecolor{cClinical}{HTML}{2563EB}   %
\definecolor{cReports}{HTML}{B45309}    %
\definecolor{cRNA}{HTML}{059669}        %
\definecolor{cMut}{HTML}{DC2626}        %
\definecolor{cCNV}{HTML}{7C3AED}        %
\definecolor{cHisto}{HTML}{4338CA}      %
\definecolor{cPool}{HTML}{0F766E}       %
\definecolor{cTeacher}{HTML}{7C3AED}    %
\definecolor{cStudent}{HTML}{0D9488}    %
\definecolor{cDoctor}{HTML}{059669}     %
\definecolor{cBrain}{HTML}{DB2777}      %
\definecolor{cGold}{HTML}{D97706}       %
\definecolor{cKey}{HTML}{EAB308}        %
\definecolor{cPatient}{HTML}{2563EB}    %

\begin{tikzpicture}[
  font=\footnotesize,
  every node/.style={align=center},
  patient/.style={rectangle, rounded corners=2.5pt, draw=cPatient!70,
    fill=cPatient!5, minimum width=20mm, minimum height=8.5mm,
    inner sep=1.5pt, thick},
  privbox/.style={rectangle, rounded corners=2.5pt, draw=cKey!80,
    fill=cKey!8, minimum width=20mm, minimum height=8.5mm,
    inner sep=1.5pt, thick},
  teacher/.style={rectangle, rounded corners=3pt, draw=cTeacher!80,
    fill=cTeacher!10, minimum width=38mm, minimum height=11mm,
    inner sep=2pt, very thick},
  student/.style={rectangle, rounded corners=3pt, draw=cStudent!80,
    fill=cStudent!10, minimum width=38mm, minimum height=11mm,
    inner sep=2pt, very thick},
  loss/.style={rectangle, rounded corners=2.5pt, draw=cGold!80,
    fill=cGold!10, minimum width=38mm, minimum height=7.5mm,
    inner sep=1.5pt},
  pred/.style={rectangle, rounded corners=2.5pt, draw=cBrain!70,
    fill=cBrain!7, minimum width=38mm, minimum height=8mm,
    inner sep=1.5pt},
  callout/.style={rectangle, rounded corners=2.5pt, draw=black!30,
    fill=yellow!14, minimum width=90mm, inner sep=2.5pt},
  mod/.style={rectangle, rounded corners=2.5pt, draw=black!45,
    fill=white, minimum width=15.5mm, minimum height=10.5mm,
    inner sep=1.5pt},
  arr/.style={-{Latex[length=2mm,width=1.6mm]}, thick, black!60},
  darr/.style={-{Latex[length=2mm,width=1.6mm]}, very thick, cTeacher!80},
  sarr/.style={-{Latex[length=2mm,width=1.6mm]}, very thick, cStudent!80},
  privarr/.style={-{Latex[length=2mm,width=1.6mm]}, dashed, thick, cKey!85},
]

\coordinate (poolctr) at (0, 0);

\node[mod, xshift=-58mm, yshift=-1mm,
      draw=cClinical!60, fill=cClinical!5]
  at (poolctr)
  (m1) {{\normalsize\color{cClinical}\faNotesMedical}\\[-1pt]
        \tiny Test 1\\[-2pt] \tiny free};
\node[mod, right=0.8mm of m1, draw=cReports!60, fill=cReports!6]
  (m2) {{\normalsize\color{cReports}\faFileMedical}\\[-1pt]
        \tiny Test 2\\[-2pt] \tiny\$1.5};
\node[mod, right=0.8mm of m2, draw=cRNA!60, fill=cRNA!6]
  (m3) {{\normalsize\color{cRNA}\faDna}\\[-1pt]
        \tiny Test 3\\[-2pt] \tiny\$3.0};
\node[mod, right=0.8mm of m3, draw=cMut!60, fill=cMut!6]
  (m4) {{\normalsize\color{cMut}\faVial}\\[-1pt]
        \tiny Test 4\\[-2pt] \tiny\$2.0};
\node[mod, right=0.8mm of m4, draw=cCNV!60, fill=cCNV!6]
  (m5) {{\normalsize\color{cCNV}\faChartArea}\\[-1pt]
        \tiny Test 5\\[-2pt] \tiny\$2.0};
\node[mod, right=0.8mm of m5, draw=cHisto!60, fill=cHisto!6]
  (m6) {{\normalsize\color{cHisto}\faMicroscope}\\[-1pt]
        \tiny Test 6\\[-2pt] \tiny\$4.0};

\begin{scope}[on background layer]
\node[rounded corners=5pt, draw=cPool!75, very thick,
      fill=cPool!4,
      inner xsep=3.5mm, inner ysep=3mm,
      fit=(m1)(m6)] (poolbox) {};
\end{scope}
\draw[cPool!85, line width=1.4mm, line cap=round]
  ($(poolbox.north west)+(1.6mm,-3mm)$) --
  ($(poolbox.south west)+(1.6mm,3mm)$);
\node[fill=white, draw=cPool!75, very thick, rounded corners=3pt,
      inner xsep=3pt, inner ysep=1.5pt,
      font=\bfseries\footnotesize, text=cPool!80!black]
  at (poolbox.north)
  (poolhdr)
  {{\color{cPool}\faStethoscope}\ Modality pool: available diagnostic tests};

\coordinate (trainorigin) at (-38mm, -27mm);

\node[font=\bfseries\footnotesize, text=cTeacher!80!black]
  at ($(trainorigin)+(0,11mm)$) (trainhdr)
  {{\color{cTeacher}\faChalkboardTeacher}\ Training time (offline)};

\node[patient] at (trainorigin) (train-x)
  {{\normalsize\color{cPatient}\faUser}\ Patient $i$\\[-1pt]
   \tiny observed $\bx_S$};
\node[privbox, right=4mm of train-x] (train-priv)
  {{\normalsize\color{cKey}\faKey}\ \textbf{Privileged}\\[-1pt]
   \tiny label $y_i$,\ peek $x^{(a)}$};

\node[teacher, below=5mm of train-x, xshift=14mm] (teacher)
  {{\normalsize\color{cTeacher}\faUserTie}\ %
   \textbf{\color{cTeacher!85!black}Privileged Oracle}\\[-1pt]
   \tiny myopic Brier oracle -- an \emph{algorithm}};

\node[loss, below=3mm of teacher] (softtarget)
  {{\color{cGold}\faTrophy}\ soft target $q_S^{(i)}(a)$};

\node[student, below=3mm of softtarget] (student1)
  {{\normalsize\color{cStudent}\faUserGraduate}\ %
   \textbf{\color{cStudent!85!black}Student} $\pi_\phi$\\[-1pt]
   \tiny mask-conditioned MLP (deployable)};

\draw[privarr] (train-x.south) -- (teacher.north-|train-x);
\draw[privarr] (train-priv.south) -- (teacher.north-|train-priv);
\draw[darr] (teacher) -- (softtarget);
\draw[darr] (softtarget) --
  node[right, font=\tiny\itshape, text=cTeacher!80!black]{\ soft-KL}
  (student1);

\draw[dashed, black!30, thick]
  (0mm, -19mm) -- (0mm, -74mm);

\coordinate (deployorigin) at (38mm, -27mm);

\node[font=\bfseries\footnotesize, text=cStudent!80!black]
  at ($(deployorigin)+(0,11mm)$) (deployhdr)
  {{\color{cStudent}\faStethoscope}\ Deployment (online)};

\node[patient, fill=cStudent!5, draw=cStudent!60] at (deployorigin) (deploy-x)
  {{\normalsize\color{cStudent}\faUser}\ New patient\\[-1pt]
   \tiny observed $\bx_{q,S}$};

\node[student, below=3.5mm of deploy-x] (student2)
  {{\normalsize\color{cStudent}\faUserGraduate}\ %
   \textbf{\color{cStudent!85!black}Student} $\pi_\phi$\\[-1pt]
   \tiny (same weights, no teacher, no label)};

\node[patient, below=3mm of student2, fill=cDoctor!7, draw=cDoctor!65,
  minimum width=38mm]
  (nextmod)
  {{\normalsize\color{cDoctor}\faUserMd}\ %
   \textbf{\color{cDoctor!80!black}Acquire next modality} $a^\star$};

\node[pred, below=3mm of nextmod] (pred)
  {{\color{cBrain}\faBrain}\ %
   predictor $f_\theta(\bx_{q,S\cup\{a^\star\}}) \to \hat y$};

\draw[arr] (deploy-x) -- (student2);
\draw[sarr] (student2) -- (nextmod);
\draw[sarr, dashed] (nextmod.east) -- ++(4mm,0) --
  node[right, font=\tiny\itshape, text=cStudent!80!black]
    {\ loop\\ \ until\\ \ budget}
  ($(student2.east)+(4mm,0)$) -- (student2.east);
\draw[arr] (nextmod) -- (pred);

\coordinate (poolexit) at (poolbox.south east);
\coordinate (rightlane) at ($(nextmod.east)+(11mm,0)$);
\draw[-{Latex[length=2mm,width=1.6mm]}, dashed, thick, cDoctor!60,
      rounded corners=3pt]
  (poolexit) -- ++(0,-2mm) -| (rightlane) -- (rightlane |- nextmod.east)
  -- (nextmod.east);

\end{tikzpicture}
\caption{\textbf{\ourmethod\ pipeline as a clinical diagnostic workflow.}
\emph{The privileged oracle runs only at training time; the student is the only
component that deploys.} \emph{Modality pool (top):} the acquisition policy
chooses among distinct diagnostic tests with heterogeneous per-test costs. \emph{Training time
(left):} the privileged oracle is a myopic, deterministic Brier algorithm with
access to the label $y_i$ and the unobserved candidate value $x^{(a)}$; for each
sampled $(x_i, S)$ it ranks the candidates, and the deployable student policy
$\pi_\phi(a \mid \bx_S, \bmm_S)$ is trained to reproduce that ranking by soft-KL
distillation under a random-mask prior. \emph{Deployment (right):} the student
selects the next modality under the remaining budget, with no oracle
recomputation and no access to $y$ or $x^{(a)}$. The pretrained predictor
$f_\theta$ produces the final diagnosis.}
\label{fig:pipeline}
\end{figure*}

In many applications, obtaining the feature values of a data
instance is not free. A clinical diagnosis draws on tests that
differ widely in monetary cost, turnaround time, and invasiveness \citep{ma2019eddi};
 discovering a new material or molecule requires costly wet-lab experiments or simulations, each
consuming reagents, compute, or instrument time
\citep{shahriari2016bayesopt, meng2026pde, ding2024learning, ding2026texttwintranslation};  and a sensing or robotic platform must spend power, time, or motion for each measurement
\citep{krause2008sensor}.
Therefore, acquiring the full feature set for every instance is often infeasible, and even when it is possible the cost rarely justifies the marginal
gain. Active Feature Acquisition (AFA) \citep{ma2019eddi} addresses this by
learning to select, for each instance, a small and informative subset of
features to acquire, maximizing predictive performance under a fixed acquisition
budget.

Progress in the field has come from increasingly sophisticated
ways of constructing an \emph{acquisition objective}: an information
reward under a partial generative model \citep{ma2019eddi},
conditional mutual information estimated by a learned network
\citep{covert2023gdfs, gadgil2024dime}, a non-parametric
acquisition oracle \citep{valancius2023aaco}, or long-horizon
return optimized by reinforcement learning
\citep{shim2018jafa, zannone2019odin}. Despite the algorithmic
diversity, all of these methods share a structural property: the
supervision signal is \emph{self-constructed} -- these methods both
define the acquisition objective and optimize toward it, and the acquisition policy is never supervised by an external teacher. This naturally raises a question. At training time we already have access to information the deployed policy never sees: the true label, and the actual values of the candidates that have not been acquired, which are exactly the ingredients a myopic acquisition oracle needs to rank candidates in closed form. \textit{If that supervision is available for free, why spend the algorithmic effort designing yet another surrogate?}

We instead construct the acquisition teacher \emph{explicitly}, as a deterministic
algorithm that scores each candidate by the error the downstream predictor would
incur were it acquired. We then train a deployable student to reproduce the teacher's ranking. Distilling a teacher's ranking into a student is a \emph{known} method, but prior work splits into two families that our setting sits between. On the one hand, in capacity distillation the teacher is a learned, larger model and the students exist to cut inference cost as in ranking distillation for recommendation ~\citep{tang2018ranking, reddi2021rankdistil} and reranker compression for large language models ~\citep{sun2023chatgpt}. On the other hand, in privileged-information distillation the teacher can see inputs the students cannot access at test time and the student aims to close that information gap rather than to save compute ~\citep{lopez2015unifying}. \ourmethod\ belongs to the second family and sharpens it in two aspects to tailor for AFA settings. First, our teacher is not a learned model but a closed form oracle computed from the training labels and the unacquired feature values, so there is no teacher to train and thus no teacher approximation error. Second, the distillation object is a \emph{budget-constrained} acquisition ranking, so the student is re-queried at each partial-observation state and only its top candidate is acquired per step. The resulting student, \ourmethod, is a mask-conditioned network trained by soft-KL under a random-mask prior, and it never sees the label or the
unacquired feature values at test time (Figure~\ref{fig:pipeline}).

Across multimodal medical diagnosis and the AFABench benchmarks~\citep{schutz2025afabench}, six datasets in
total, this approach matches or outperforms specialized utility-learning
methods across the budget range, and on
the medical cohort recovers $90\%$ of the attainable accuracy gain at
under a quarter of the acquisition cost. Making the
teacher an explicit algorithmic object also makes it analyzable. Once the
teacher is a concrete mapping from observed state to ranked action, we can ask
questions the AFA literature could not previously address directly: how much of the
teacher's behavior is recoverable by a student that sees only what is visible at
deployment, how student quality depends on the teacher it distils, and which
alternative explanations for the residual oracle gap can be excluded by
controlled ablations.

\subsection*{Contributions.}

\begin{enumerate}
\item We reframe AFA as learning from an explicit privileged
acquisition teacher rather than from a self-constructed utility
surrogate (Section~\ref{sec:related}, formalized in
Section~\ref{sec:method}).
\item We introduce \ourmethod, a deployable student policy trained
by soft-target distillation from a myopic Brier oracle under a
random-mask prior (Section~\ref{sec:method}), and show it matches or
outperforms specialised utility learners across six datasets
(Section~\ref{sec:experiments}).
\item We use the explicit teacher as an instrument to study the
AFA problem itself, reporting a teacher-quality sweep
(Section~\ref{sec:teacher-quality}), an empirical
teacher-learnability diagnostic, and controlled ablations that
exclude architecture, predictor, cost scheme, and generalization as
explanations for the observed saturation
(Section~\ref{sec:diagnostic}).
\end{enumerate}

\section{Related Work}
\label{sec:related}

\paragraph{Active feature acquisition (AFA).} AFA methods generate their acquisition supervision internally and differ mainly in how. One line of work optimizes a self-constructed utility (as stated in Section~\ref{sec:intro}): EDDI~\citep{ma2019eddi} uses an information-theoretic acquisition reward under a partial VAE, \citet{covert2023gdfs} amortize the greedy conditonal-mutual-information policy, and DIME ~\citep{gadgil2024dime} estimates the conditional mutual information itself. AACO ~\citep{valancius2023aaco} applies a nonparametric acquisition-conditioned oracle at inference, whereas we distill an offline privileged oracle into a separate deployable policy. A second line of work formulates acquisition as sequential decision-making problem and optimizes a policy by RL~\citep{shim2018jafa, zannone2019odin, kachuee2018ol, song2025viewselect}. These methods differ in what they optimize, utility versus return, but not in where the signal comes from: the target is designed, not distilled from a teacher. A recent benchmark further finds that the RL-based methods often train unstably and do not consistently outperform simpler greedy or oracle-based alternatives~\citep{schutz2025afabench}.

\paragraph{Learning from teachers.} A separate line of work learns from an \emph{existing} supervisor. Imitation learning and behavior cloning
\citep{pomerleau1988alvinn, ross2011dagger, osa2018imitation} distil an
expert's demonstrations; learning using privileged information
\citep{vapnik2015privileged, vapnik2009new} transfers a
training-time-only signal to the deployed model; and knowledge distillation
\citep{hinton2015distilling, buciluua2006model} matches a student to a
pretrained teacher model. In all cases, these works study how to
\emph{transfer} knowledge from an existing teacher, rather than whether such
a teacher can be \emph{constructed} in the first place.

\paragraph{Prediction under missing modalities.} Predicting from an
arbitrary observed subset is complementary to AFA: mask-aware predictors trained under random feature masking produce a label from whatever has been acquired ~\citep{schutz2025afabench, gadgil2024dime, covert2023gdfs}. We take such a predictor as given and study only which modalities to acquire for it.

\section{Problem Setup}
\label{sec:setup}

\subsection{Setting}
\label{sec:setting}

Let $x = (x^{(1)}, \ldots, x^{(K)}) \in \R^d$ be a feature vector
partitioned into $K$ \emph{modalities}, the units at which acquisition
happens. In our motivating medical cohort a modality is a distinct
diagnostic test (a clinical, transcriptomic, or histology panel), each
separately orderable; in the tabular benchmarks a modality is an
individual feature. Modality $k$ carries an acquisition cost
$c_k \ge 0$, and we write $c(S) = \sum_{k \in S} c_k$ for the cost of a
subset $S \subseteq [K]$. A designated set $\mathrm{FREE} \subseteq [K]$ is available at
no cost as $c_k = 0$ for $k \in \mathrm{FREE}$. In the medical setting these are the
demographics recorded at intake. A subset $S \subseteq \{1,\ldots,K\}$
of already-observed modalities induces a masked observation $\bx_S \in \R^d$
(values outside $S$ are set to zero) and a binary mask
$\bmm_S \in \{0,1\}^K$ indicating $S$. 

At each step $t$ an acquisition policy $\pi$ observes
$(\bx_{S_t}, \bmm_{S_t})$ and selects a new modality $a_t \notin S_t$,
giving $S_{t+1} = S_t \cup \{a_t\}$ and cumulative cost
$c(S_{t+1}) = c(S_t) + c_{a_t}$. The acquisition process proceeds until the budget $B$
is exhausted. A downstream classifier
$f_\theta(\bx_S, \bmm_S) \to \hat p(y \mid \bx_S) \in \Delta^{C-1}$ maps
any observed subset to a class distribution, with prediction
$\hat y_\theta(\bx_S) = \Argmax f_\theta(\bx_S, \bmm_S)$. We train $f_\theta$
once, on randomly masked subsets, and hold it fixed across every acquisition
policy we compare. Holding $f_\theta$ fixed makes the downstream
predictor a fixed environment rather than a moving target, so that differences
in accuracy are \emph{attributable} to the acquisition policy alone.

\subsection{Learning objective}
\label{sec:objective}

Call a policy \emph{deployable} if it conditions only on the observed
$(\bx_S, \bmm_S)$, that is, never on the label $y$ or on the values of unacquired
modalities. The subset a deployable policy
$\pi$ acquires on input $\bx$ under budget $B$ is 
$S_\pi \equiv S_\pi(\bx, B)$. Our objective is the budgeted downstream accuracy
\begin{equation}
\label{eq:objective}
\max_{\pi\;\text{deployable}} \;
  \mathbb{E}_{(\bx,y)}\!\left[\mathbf{1}[\hat y_\theta(\bx_{S_\pi}) = y]\right]
\quad\text{s.t.}\quad c(S_\pi) \le B ,
\end{equation}
evaluated across the range of budgets $B$ rather than at a single operating point. This objective rejects direct optimization in two ways. First, it is
non-differentiable, as $\pi$ enters only through the discrete set $S_\pi$ and accuracy is an indicator. Second, it is not a property of $\pi$ alone: whether a subset was worth acquiring depends on $f_{\theta}$, and how far down the acquisition order the policy is evaluated depends on $B$. Prior work (§\ref{sec:related}) optimizes this objective through
RL or a hand-built acquisition utility. We instead exploit the fact that a good acquisition order is easy to identify with training-time information: given $y$ and the full $\bx$, one can compute directly how much each candidate modality would change the downstream error under $f_\theta$. What is hard is identifying it \emph{deployably}, from $\bx_S$ alone. We therefore compute the signal offline on the training data and learn a deployable policy that imitates it, as described in the next section.

\section{Method: \ourmethod}
\label{sec:method}

The objective in Equation~\ref{eq:objective} is hard to optimize but easy to evaluate offline: given a label and a full feature vector, the effect of each candidate modality on the downstream error can be measured directly, even though neither the label nor the unacquired values is available at deployment.

\paragraph{Our method.} We treat the offline computation as a \emph{teacher} and learn a deployable \emph{student} that imitates it.
On training data, we score every unacquired modality by the error the shared predictor would incur if that modality were added, and turn the scores into a ranking over candidates. We then train a policy that sees only the observed subset and its mask to reproduce that ranking. The teacher is discarded after training and the student is deployable as-is.

\subsection{The teacher}
\label{sec:teacher}
For a training
pair $(x_i, y_i)$ and observed subset $S$, the teacher scores each candidate $a \notin S$ by the Brier score~\citep{glenn1950verification} the predictor would incur after acquiring it:
\begin{equation}
\label{eq:oracle}
b_a^{(i,S)}
  = \left\| f_\theta\!\left(\bx^{(i)}_{S \cup \{a\}}, \bmm_{S \cup \{a\}}\right)
      - \mathbf{e}_{y_i} \right\|^2 ,
\qquad a \notin S ,
\end{equation} 
We score with a strictly proper rule rather than 0-1 loss so that the teacher ranks candidates by the quality of the full predictive distribution rather than argmax alone~\citep{gneiting2007strictly}. Boundedness also keeps a single catastrophic candidate from saturating the softmax target. Since the pre-acquisition error is common to all candidates, minimizing $b_a$ is the same as maximizing the error \emph{reduction} and the preferred candidate is $\Argmin_{a \notin S} b_a^{(i,S)}$. Following the terminology of \citet{vapnik2015privileged}, the teacher is \emph{privileged} in two ways, both of which disqualify at deployment: it reads the label $y_{i}$, and it reads the value $x^{(a)}$ of a modality it has not paid for.

\paragraph{Why a myopic teacher?} A natural alternative is a finite-horizon oracle that plans several acquisitions ahead. We use the one step version instead, for two reasons. First, it is well-defined given nothing but the trained $f_\theta$, whereas a planning oracle introduces its own search hyperparameters and its own approximation error. Second, on our datasets the gains from non-myopic planning are limited, consistent with the benchmark finding that this benefit is strongly dataset-dependent~\citep{schutz2025afabench}. We revisit this choice in Appendix~\ref{app:oracle-choice}.

\paragraph{Why this makes distillation easy.} Two properties of teacher matter for downstream accuracy. First, it is an algorithm rather than a fitted model: given $(x_i, y_i, S)$ and $f_\theta$, its scores are determined. There is no teacher to train on and no teacher approximation error between the student and the objective, in contrast to distilling a trained network~\citep{hinton2015distilling, vapnik2015privileged}. Second, the $f_\theta$ inside the teacher is the \emph{same} predictor the student passes through at deployment, so the acquisition signal is measured on the model that will actually be scored. No auxiliary per-mask classifier is trained.

\subsection{The student}
\label{sec:model}

The student $\pi_\phi(a \mid \bx_S, \bmm_S)$ conditions only on the observed subset and its mask, which makes it deployable by construction. It is a small mask-conditioned MLP mapping $[\bmm_S \odot \bx, \bmm_S]$ through a shared trunk into an \emph{action head}, with one logit per candidate, and an auxiliary \emph{classifier head} used only during training as a multi-task regularizer~\citep{caruana1997multitask, kendall2018multitask, ding2024learning}. At deployment we take $\Argmax$ over the unobserved, affordable candidates from the action head and discard the outputs of classifier head. Training the action head is behavior cloning of the teacher ~\citep{pomerleau1988alvinn}: we minimize cross-entropy against the teacher's selection on states drawn from randomly masked subsets. Because random masking exposes the student to a broad range of reachable partial-observation states, it mitigates the covariate shift that motivates interactive imitation learning ~\citep{ross2011dagger}.  Appendix~\ref{app:implementation} details the widths, depths and parameter counts of the MLP.

\subsection{Training and deployment}
\label{sec:algo}

\paragraph{Which subsets should the student see?} Training only on subsets the teacher itself visits would expose the student to a narrow slice of the state space and leave it fragile off that trajectory, the familiar covariate shift of behavior cloning ~\citep{ross2011dagger}. We instead sample subsets from a Bernoulli-of-Bernoulli prior: draw $p_i \sim \mathcal{U}(0,1)$ and include each \emph{acquired} modality independently with probability $p_i$, always keeping the free tier so every sampled state is reachable from $S_0 = \mathrm{FREE}$. This covers every subset size at comparable rates, mimicking masked-input pretraining in language models ~\citep{devlin2019bert}.

\paragraph{What does the student match?} We convert teacher scores into a soft target via a temperature-scaled softmax over $-b_a^{(i, S)}$ and minimize $\mathcal{L} = \mathcal{L}_\text{act} + \lambda_\text{cls} \mathcal{L}_\text{cls}$, a soft cross-entropy (CE) against the teacher distribution plus label CE on the trunk. It turns out both choices matter on small datasets: distilling the full ranking rather than the argmax alone, and the auxiliary term, which supplies a dense per-sample gradient toward $p(y \mid \bx_S)$ that the sparse action target lacks (Appendix~\ref{app:training-recipe}). Algorithm~\ref{alg:train} summarises the full training procedure.

\begin{algorithm}[h]
\caption{\ourmethod\ Training.}
\label{alg:train}
\begin{algorithmic}[1]
\REQUIRE Training set $\{(x_i, y_i)\}_{i=1}^N$, pretrained predictor
$f_\theta$, temperature $T$, weight $\lambda_\text{cls}$, epochs $E$.
\STATE Initialize student $\pi_\phi$.
\FOR{$e = 1, \ldots, E$}
  \FOR{minibatch $B$}
    \FOR{each $(x_i, y_i) \in B$}
      \STATE Sample $p_i \sim \mathcal{U}(0,1)$.  
      \STATE Sample $S$ over modalities with probability $p_i$.
      \STATE Construct $(\bx_S^{(i)}, \bmm_S^{(i)})$.
      \STATE $b_a^{(i,S)} \gets \big\|f_\theta(\bx^{(i)}_{S \cup \{a\}}, \bmm_{S \cup \{a\}}^{(i)}) - \mathbf{e}_{y_i}\big\|^2$ $\forall a \notin S$.
      \STATE $q_S^{(i)}(a) \gets \left[ \Softmax_{a'\notin S}\!\left(-b^{(i,S)}/T\right) \right]_a$.
      \STATE $(z^\text{act}_i, z^\text{cls}_i) \gets \pi_\phi(\bx^{(i)}_S, \bmm_S^{(i)})$
      \STATE $\mathcal{L}^\text{act}_i \gets -\sum_{a \notin S} q_S^{(i)}(a)\,\log\big[\Softmax(z^\text{act}_i)\big]_a$
      \STATE $\mathcal{L}^\text{cls}_i \gets -\log\big[\Softmax(z^\text{cls}_i)\big]_{y_i}$
    \ENDFOR
    \STATE Update $\phi$ with
    $\nabla_\phi \frac{1}{|B|}\sum_{i\in B} (\mathcal{L}^\text{act}_i +
    \lambda_\text{cls} \mathcal{L}^\text{cls}_i)$.
  \ENDFOR
\ENDFOR
\RETURN $\pi_\phi$.
\end{algorithmic}
\end{algorithm}

\paragraph{Deployment.} 
Given a query $\bx_q$ and budget $B$, the student starts from $\mathrm{FREE}$ and repeatedly takes the highest-scoring candidate that is unobserved and still affordable, stopping when none remains, then predicts through $f_\theta$. This process uses no label, no unacquired feature, and no teacher. We include no \texttt{STOP} action: the budget is per-instance rather than pooled, so stopping early cannot buy accuracy elsewhere, and the full budget-accuracy curve we report already exposes every operating point a learned \texttt{STOP} could select. Algorithm~\ref{alg:deploy} details the full deployment process.

\begin{algorithm}[h]
\caption{\ourmethod\ Deployment.}
\label{alg:deploy}
\begin{algorithmic}[1]
\REQUIRE Query $\bx_q$, trained student $\pi_\phi$, pre-trained
predictor $f_\theta$, budget $B$, per-modality costs
$\{c_a\}_{a=1}^K$, free tier $\mathrm{FREE}$.
\STATE $S \gets \mathrm{FREE}$, \ $c \gets 0$.
\WHILE{$c < B$ \AND $\exists a \notin S : c + c_a \le B$}
  \STATE $z \gets \pi_\phi(\bx_{q,S}, \bmm_S)$.
  \STATE $z_a \gets -\infty$ for every $a \in S$ or with
  $c + c_a > B$.
  \STATE $a^\star \gets \Argmax_a z_a$.
  \STATE $S \gets S \cup \{a^\star\}$;\ \ $c \gets c + c_{a^\star}$.
\ENDWHILE
\STATE $\hat y \gets \Argmax f_\theta(\bx_{q,S}, \bmm_S)$.
\RETURN $\hat y$,\ $S$,\ $c$.
\end{algorithmic}
\end{algorithm}

\paragraph{Cost.} The expensive step, scoring every candidate through $f_\theta$ to build the teacher targets, run once offline during distillation. At deployment \ourmethod\ costs at most $K$ student passes plus the fixed predictor $f_\theta$, matching policy-based AFA methods and much cheaper than baselines that re-run a generative estimator at each step (cf. Appendix~\ref{app:complexity}).
\section{Main Results}
\label{sec:experiments}

\subsection{Experimental setup}
\label{sec:exp-setup}

\paragraph{Datasets.} The first dataset, TCGA-Glioma \citep{heath2021gdc}, serves as our motivating dataset: 
a real-world multimodal medical cohort in which each acquisition unit corresponds to a distinct modality with clinically-informed acquisition costs assigned by domain experts. The remaining datasets are not meant
as real-world demonstrations; they test whether it is \emph{robust} across
granularities and domains. UCI-Heart \citep{detrano1989heart} is a clinical diagnosis task, but its
acquisition units are \emph{individual features} rather than separately
orderable tests; UCI-Adult \citep{kohavi1996adult}, Cube \citep{shim2018jafa, schutz2025afabench}, and UCI-MiniBooNE \citep{roe2005miniboone} are census, synthetic,
and large-tabular benchmarks at that same feature granularity.
PhysioNet-2012 \citep{silva2012physionet, goldberger2000physionet} sits in between, grouping intensive-care measurements into
standard clinical panels. Unlike TCGA-Glioma, where an acquisition unit corresponds to a clinically meaningful modality, these benchmarks define acquisition at the feature level with each feature treated as an acquisition action. We provide further details on these choices of encoders, cost schedules, and dataset statistics in Appendix~\ref{app:datasets}.

\paragraph{Predictors.} We evaluate \ourmethod\ with two downstream
predictors, each held fixed across every acquisition method
(§\ref{sec:setting}). (i)~A standard \emph{late-fusion} stacking
classifier \citep{song2026multimodal}: it is well suited to multimodal
clinical diagnosis, attains the stronger predictive accuracy, and --
because it fits a separate fusion head per observed subset -- most sharply
rewards a correct acquisition, which is where our teacher-aligned policy
helps most (Appendix~\ref{app:why-late-fusion}). (ii)~The shared
\emph{mask-aware MLP} adopted by AFABench \citep{schutz2025afabench}, on
which we show \ourmethod\ is \emph{predictor-agnostic}
(Appendix~\ref{app:predictor-agnostic}).

\paragraph{Evaluation.} We report held-out accuracy across the full range
of budgets (the budget--accuracy curves of Fig.~\ref{fig:headline});
every reported number is averaged over 5 random seeds, and the medical
datasets additionally use 5-fold cross-validation. The full protocol --
per-dataset budgets, seed handling, and the baseline set -- is in
Appendix~\ref{app:datasets}.

\subsection{Main results}
\label{sec:main-results}
\label{sec:experiments-headline}
\label{sec:experiments-source-b}

\begin{figure*}[t]
\centering
\includegraphics[width=\linewidth]{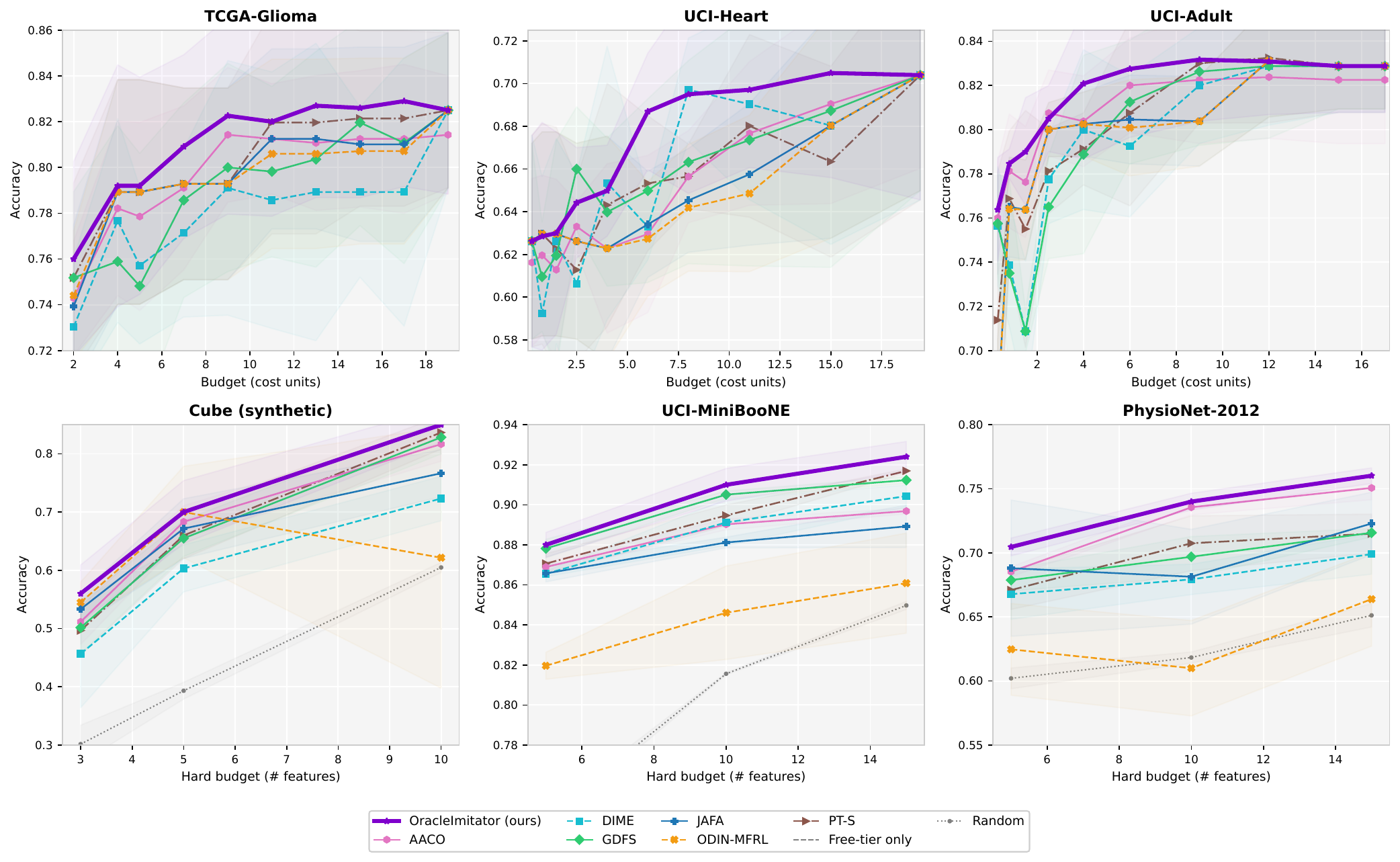}
\caption{\textbf{Performance against acquisition budget across six different datasets.}
Row 1: \emph{late-fusion} predictor on Glioma (complexity-based cost),
Heart, and Adult,
5-fold CV; shaded bands are $\pm 1$\,s.d.\ across folds. On Glioma every
method is aggregated identically as a plain mean over seeds, with no
best-of-$n$ selection. Row 2:
AFABench's shared \emph{mask-aware MLP} predictor on Cube, MiniBooNE, and
PhysioNet; shaded bands are $\pm 1$\,s.d.\ across seed-runs.}
\label{fig:headline}
\end{figure*}

Figure~\ref{fig:headline} compares \ourmethod\ with representative AFA
baselines across six datasets that span both modality-level and
feature-level acquisition. \ourmethod\ consistently matches or exceeds the
strongest competing methods across the evaluated budget range. This holds
under two substantially different acquisition granularities: clinically
meaningful modalities with heterogeneous costs in TCGA-Glioma (top row),
and per-feature acquisition under the standard AFABench protocol (bottom
row), to which \ourmethod\ transfers without modification. The advantage is most pronounced at intermediate budgets, where acquisition decisions matter most:  on TCGA-Glioma \ourmethod\ recovers $90\%$ of the full-acquisition
accuracy gain at under a quarter of the total cost, reaching competitive
accuracy with fewer acquired modalities rather than only after nearly full
acquisition. The separation between methods is generally wider under the late-fusion
predictor (top row) than under the shared mask-aware MLP (bottom row). We
revisit this in Appendix~\ref{app:why-late-fusion}, where we show that
late fusion amplifies the effect of correct acquisition decisions because
each observed subset is served by its own per-subset predictor. Beyond the aggregate accuracy gains,  \ourmethod\ learns patient-specific acquisition strategies rather than a fixed acquisition order. Patients share a common low-cost opening, but their acquisition trajectories increasingly diverge as additional budget becomes available (Appendix ~\ref{app:trajectories}). These results establish that explicit teacher supervision is competitive
across both acquisition granularities and predictor families.

\subsection{Robustness and scalability}
\label{sec:ablations}

\paragraph{Robustness.} The main conclusions are robust to both the
acquisition \textit{cost definition} and the \textit{downstream predictor}. Replacing the
complexity-based Glioma schedule with either a unit-cost schedule or a
fixed-seed permutation of the same costs preserves the method ordering and
\ourmethod's competitiveness (Appendix~\ref{app:cost-robustness}); under
the permutation -- which makes the cheapest, most informative units the
most expensive -- the strongest baseline drops from second to last while
\ourmethod\ remains first. Swapping the late-fusion predictor for a shared
mask-aware MLP likewise preserves \ourmethod's standing in a controlled
head-to-head (Appendix~\ref{app:predictor-agnostic}).

\paragraph{Scalability.} The offline teacher computation scales to large
candidate sets through a simple candidate-pruning heuristic: a $k$NN proxy
ranks candidates and the exact Brier is computed only on the top $M$.
Restricting evaluation to $M{=}5$ costs at most $0.7$\,pp of student
accuracy at $K{=}12$ while cutting teacher computation by roughly
$2.4\times$, so the teacher scales with $K$ as an engineering knob without
changing the student (Table~\ref{tab:teacher-pruning},
Appendix~\ref{app:teacher-pruning}).

\subsection{Teacher ablation}
\label{sec:teacher-quality}

\ourmethod\ depends on a privileged teacher, so a natural design question is how much the choice of teacher actually matters. We retrain the imitator under three teacher settings while holding
architecture, random-mask augmentation, and hyper-parameters
fixed:
\begin{itemize}
\item \textbf{Random teacher.} Soft target uniform over un-observed
candidates at every $(x_i, S)$.
\item \textbf{Entropy teacher.} Teacher score is the
entropy of $P(y \mid x^{(a)}, \bx_S)$; the teacher peeks at
$x^{(a)}$ but does not know $y$. 
\item \textbf{Brier teacher (\ourmethod).} Teacher score is
$\|f_\theta(\bx_{S \cup \{a\}}) - \mathbf{e}_y\|^2$: both $y$ and
$x^{(a)}$ are used at training time.
\end{itemize}

\begin{table}[h]
\centering\small
\caption{Teacher-quality sweep. Mean-across-budgets accuracy, 5-fold CV, mean over seeds per fold.}
\label{tab:teacher-quality}
{\setlength{\tabcolsep}{4.5pt}%
\begin{tabular}{lccc}
\toprule
Dataset & Random & Entropy ($y$-blind) & Brier (main) \\
\midrule
Heart      & 0.633 $\pm$ 0.014 & 0.646 $\pm$ 0.035 & \textbf{0.655 $\pm$ 0.039} \\
Glioma     & 0.767 $\pm$ 0.039 & 0.800 $\pm$ 0.041 & \textbf{0.810 $\pm$ 0.039} \\
Adult      & 0.771 $\pm$ 0.019 & 0.790 $\pm$ 0.023 & \textbf{0.801 $\pm$ 0.024} \\
\bottomrule
\end{tabular}}
\end{table}

The three teachers form a quality ladder on every dataset:
random~$<$~entropy~$<$~Brier. The random-teacher student is
$2$--$5$\,pp below the Brier-teacher student on all three datasets --
non-trivially above chance (it still learns to predict $y$ from
$\bx_S$ via the auxiliary CE head and still benefits from the
random-mask augmentation), but noticeably below the informed
student. The entropy teacher (which does not use $y$) closes most of that gap and sits roughly $1$ pp below the Brier teacher. 

\paragraph{Implication.} These results justify the choice of the Brier teacher in the main method: better supervisory signals consistently produce better acquisition policies. At the same time, the fact that the entropy teacher nearly matches the Brier teacher suggests that much of the useful supervisory signal is already encoded in the observed-feature geometry, with privileged labels providing only a relatively small additional benefit. 

\section{Understanding the Privileged Teacher}
\label{sec:diagnostic}

Section~\ref{sec:experiments} established that \ourmethod\ works and justified the use of a privileged teacher. Making
the teacher explicit lets us ask a question earlier utility-based AFA
methods could not formulate: \emph{how much of the teacher's behavior is
recoverable by a student that sees only the deployment-visible
observation $\bx_S$?} We estimate that recoverability
(§\ref{sec:diagnostic-identifiability}), validate the estimate
(§\ref{sec:continuity-support-body}), and discuss the result (§\ref{sec:diagnostic-ruling-out}). 

\subsection{Estimator of recoverability}
\label{sec:diagnostic-identifiability}

A student that sees only $\bx_S$ cannot recover the
teacher in general. The teacher is an explicit mapping
$(\bx_S, x^{(a)}, y) \mapsto a^\star$ while at deployment the
student sees only $\bx_S$. Consequently, multiple teacher actions may correspond to the same deployment-visible observation, so from the student's perspective the teacher induces a conditional distribution $P(a^\star \mid \bx_S)$ rather than a deterministic mapping. A deterministic student predicts the most probable teacher action for each $\bx_S$, yielding the maximum achievable teacher agreement
\begin{equation}
\label{eq:ceiling}
\sup_\pi \E\!\left[\mathbf{1}[\pi(\bx_S) = a^\star]\right]
  = \E_{\bx_S}\!\left[\max_a P(a^\star = a \mid \bx_S)\right] ,
\end{equation}
This ceiling depends only on the ambiguity of the mapping from deployment-visible observations to teacher actions, not on the student architecture or optimization.

The conditional $P(a^\star \mid \bx_S)$ is unknown, so we
estimate the right-hand side of Eq.~\ref{eq:ceiling}
non-parametrically, from the agreement of the teacher's action across a
$k$-nearest-neighbor cohort in $\bx_S$-space,
\begin{equation}
\label{eq:learn}
R(S) = \frac{1}{N}\sum_i
  \max_a \frac{1}{k}\sum_{j \in \mathcal{N}_k(i, S)}
    \mathbf{1}\!\left[a^\star_j = a\right] ,
\end{equation}
inspired by the classical use of nearest-neighbor consistency to characterize local class ambiguity \citep{cover1967nearest}; the choice of $k$, a companion entropy
diagnostic, and per-state values are in Appendix~\ref{app:extra}.
We next ask whether this estimator faithfully reflects what a trained student can achieve.

\subsection{Validating the estimator}
\label{sec:continuity-support-body}

\textbf{Can $R(S)$ be trusted?} $R(S)$ rests on a \emph{local continuity} assumption:
nearby points in $\bx_S$-space induce similar label and teacher-action
distributions. Under this assumption, a $k$NN neighborhood provides a reasonable local estimate of the conditional distribution. If this continuity fails, $R(S)$ would not estimate anything meaningful.

We probe the assumption from two complementary directions. \textbf{(C1)} measures whether nearby points in $\bx_s$ space tend to share the same label, validating the local structure of the representation. \textbf{(C2)} measures whether the soft $k$NN label distribution agrees with the predictor's output, validating that the predictor preserves this local structure. Together, they test whether the neighborhood used by $R(S)$ is both meaningful and consistent with the predictor.

Across datasets and subset states, \textbf{C1} becomes increasingly negative as additional modalities are observed, indicating increasing local label consistency. \textbf{C2} exceeds 0.8 for most non-glioma datasets at mid-to-late acquisition states, showing strong agreement between the $k$NN estimate and the predictor. Exact definitions and full per-dataset results are provided in Appendix~\ref{sec:continuity-support}.

Together, these results indicate that the local neighborhood used in
Eq.~\ref{eq:learn} captures the same conditional structure the trained
predictor exploits, supporting its use as a recoverability estimator.

\subsection{What limits recoverability?}
\label{sec:diagnostic-ruling-out}

Across datasets, estimated recoverability ranges from 0.30 to 0.53. Even under an optimal deterministic student, only around half of teacher decisions are recoverable from deployment-visible observations, and substantially fewer on the hardest datasets. The question is whether this gap reflects an intrinsic information limit or could be reduced by better modelling.

Across four controlled ablations (student architecture, downstream
predictor, cost scheme, and student training) the method ordering is
unchanged and only small quantitative differences appear: varying the
student trunk moves mean accuracy by at most
$\sim\!1$\,pp on average (Appendix~\ref{app:architecture}); swapping the
late-fusion predictor for AFABench's shared masked-MLP
(Appendix~\ref{app:predictor-agnostic}) or applying a matched unit-cost
scheme (Appendix~\ref{app:cost-scheme}) preserves the ordering; and an
in-sample student trained on the test fold's own labels closes only
$16$--$54\%$ of the gap (Appendix~\ref{app:mechanism},
Table~\ref{tab:insample}). 

Taken together, these results indicate that the residual gap is primarily
\emph{informational} rather than algorithmic. Once the deployment-visible
observation discards information required to distinguish teacher actions,
no deterministic student can recover it through better optimization alone.
Recoverability therefore characterizes the \emph{benchmark} rather than a
particular student: datasets differ in how much of the privileged teacher's
policy is determined by deployment-visible observations.

\section{Discussion and Future Work}
\label{sec:discussion}

Beyond providing a stronger acquisition policy, the main contribution of this paper is a different way of analyzing acquisition. Making the teacher explicit turns acquisition utility from a latent optimization target into an observable supervision signal that can be analyzed, compared across designs, and distilled into a deployable policy. The competitive empirical performance also supports using explicit teachers as an analysis tool.

\paragraph{Future work.}
\begin{itemize}
\item \textbf{Beyond empirical recoverability.} Our recoverability analysis
is empirical.
Characterizing recoverability theoretically or relating it to an
information-theoretic quantity would turn a per-dataset diagnostic into a
sharper, model-independent statement about what any deployable policy can
achieve.
\item \textbf{Beyond the current teacher.} The myopic Brier oracle is only
one teacher among many. Trajectory-level, cost-sensitive, task-aware, or
counterfactual teachers are equally constructible from training-time
information, and each defines a different, analyzable supervision signal for the
same deployable policy. In this framework the teacher, rather than the student, is the primary object of design.
\item \textbf{Beyond AFA.} Many learning problems make decisions under
limited information while richer supervision is available only during
training. Whether explicit teachers can be constructed and distilled
into deployable policies in settings such as active learning
\citep{sener2018core}, experimental design \citep{shahriari2016bayesopt},
continual learning \citep{meng2026transfer}, or online experimentation
\citep{li2026incentive} is an open question.
\end{itemize}

\bibliography{references}

\appendix
\section*{Appendix}
\section{Method Details}
\label{app:implementation}

\subsection{Choice of the myopic Brier oracle}
\label{app:oracle-choice}

The teacher of Eq.~\ref{eq:oracle} makes two design choices.

\paragraph{Myopic scoring.} A finite-horizon MDP oracle would require
modeling acquisition dynamics and is substantially more expensive,
whereas the myopic oracle is determined by the trained predictor
$f_\theta$ alone. AFABench finds that the benefit of non-myopic planning is strongly dataset-dependent and often limited on its real-world benchmarks, although it can be substantial on datasets constructed to require lookahead \citep[Sec.~4]{schutz2025afabench},
consistent with our findings in Section~\ref{sec:diagnostic}. . We therefore use the myopic teacher as the simplest fully specified privileged oracle and leave trajectory-level teachers to future work.

\paragraph{Brier utility.} We use the Brier score because it is a strictly
proper scoring rule that rewards improvements in the full predictive
distribution rather than only the top-1 prediction, providing a smooth
distillation target.

\subsection{Hyperparameters}

Unless otherwise noted, all imitator experiments use the following
hyperparameters:

\begin{center}
\begin{tabular}{ll}
\toprule
Hyperparameter & Value \\
\midrule
Hidden width $H$        & 256 \\
Hidden depth $L$        & 4 \\
Dropout                 & 0.1 \\
Batch size              & 64 \\
Learning rate           & $10^{-3}$ (Adam) \\
Epochs                  & 50 \\
Teacher temperature $T$ & 0.5 \\
Classifier-head weight $\lambda_\text{cls}$ & 0.5 \\
Teacher subsampling cap & 2000 patients \\
Seeds per (dataset, fold) & 9 \\
Top-$k$ selection       & 3 \\
\bottomrule
\end{tabular}
\end{center}

For AFABench experiments (Sec.~\ref{sec:experiments-source-b}), $H = 256$, $L = 4$
are kept identical; the input dimension increases with the number of
features $D$.

\subsection{Computing infrastructure.} 
All teacher generation, student
training, and deployment experiments run on CPU: each job is allocated
2~CPU cores and 12--16\,GB of RAM on a Slurm cluster with Intel Xeon Silver
4214 (2.20\,GHz) nodes, under Ubuntu 20.04.6 LTS (Linux kernel 5.4). No GPU
is required to reproduce any experiment in the paper. The software stack is Python 3.10.17 with
PyTorch 2.9.1, scikit-learn 1.7.2, NumPy 1.26.4, SciPy 1.15.3, and pandas
2.3.3.

\section{Experimental Details}
\label{app:datasets}

\subsection{Datasets}

Table~\ref{tab:datasets-a} summarizes the six datasets used throughout
the paper: the number of acquisition units $K$, classes $C$, and samples
$N$. TCGA-Glioma is the motivating multimodal cohort, and the only one in
which an acquisition unit is a separately orderable diagnostic test
(§\ref{sec:exp-setup}). In the remaining datasets a unit is an individual
feature (except PhysioNet-2012, where units are standard clinical panels), 
and they serve to test robustness across granularities and domains
rather than to motivate the method.

\begin{table}[h]
\centering
\small
\setlength{\tabcolsep}{1mm}
\caption{Datasets. $K$ is the number of acquisition units
(including the free tier); $C$ is the number of classes; $N$ is
the number of samples used in 5-fold CV. The Cube synthetic
task is from \citet{shim2018jafa}.}
\label{tab:datasets-a}
\begin{tabular}{lcccl}
\toprule
Dataset & $K$ & $C$ & $N$ & Source \\
\midrule
Glioma     & 7  & 3 & 560  & TCGA-LGG+GBM \\
Heart      & 13 & 3 & 297  & UCI Cleveland \\
Adult      & 14 & 2 & 800  & UCI Census \\
Cube       & 12 & 8 & 1000 & Synthetic \\
MiniBooNE  & 12 & 2 & 2000 & UCI MiniBooNE \\
PhysioNet  & 10 & 2 & 1108 & PhysioNet 2012 \\
\bottomrule
\end{tabular}
\end{table}

\paragraph{Comparability to published UCI baselines.} Our Heart setup differs from the standard UCI protocol by using a three-class severity target rather than binary diagnosis, making the task substantially harder; absolute accuracies are therefore not directly comparable to commonly reported binary results. Adult is class-balanced and subsampled, so its absolute accuracy is likewise not directly comparable to the standard imbalanced benchmark.

\paragraph{Glioma: the seven acquisition units.}
The Glioma cohort exposes seven clinically distinct acquisition
units. Every acquisition action corresponds to a single
independently obtainable TCGA data source, taken as released. We use the
precomputed modality embeddings released by \citet{song2026multimodal};
Table~\ref{tab:modality-dims-glioma} lists the underlying encoder for each
unit. In brief, whole-slide images are embedded with the UNI2-h pathology
vision transformer\footnote{\url{https://huggingface.co/MahmoodLab/UNI2-h}}
\citep{chen2023general}, free-text pathology reports with a domain-adapted LLM
\citep{labrak2024biomistral}, and mutation and copy-number profiles with
Gene2Vec; each embedding is then reduced to PCA-32 per unit.

\begin{table}[h]
\centering
\small
\setlength{\tabcolsep}{1mm}
\caption{Per-unit embedding dimensions in the Glioma setup.}
\label{tab:modality-dims-glioma}
\begin{tabular}{lcc}
\toprule
Acquisition unit & Raw dim & Encoder \\
\midrule
clinical    & 5    & age, sex, KPS, MMSE, ECOG \\
reports     & 4096 & BioMistral-7B \\
mirna       & 750  & Raw miRNA counts \\
cnv         & 200  & Gene2Vec \\
methylation & 64   & Full methylation array \\
mutations   & 200  & Whole-exome calls via Gene2Vec \\
histology   & 1536 & UNI2-h \\
\bottomrule
\end{tabular}
\end{table}

\subsection{Cost schedules}

\begin{table}[h]
\centering
\small
\caption{Cost schedule for Glioma. Costs reflect real acquisition
complexity informed by TCGA experts. \emph{Unit} assigns
cost 1 to every paid unit. Free tier = clinical.}
\label{tab:costs-glioma}
\begin{tabular}{lcc}
\toprule
Acquisition unit & Complexity (main) & Unit \\
\midrule
clinical    & 0.0 & 0.0 \\
reports     & 2.0 & 1.0 \\
histology   & 2.0 & 1.0 \\
mutations   & 3.0 & 1.0 \\
mirna       & 4.0 & 1.0 \\
methylation & 4.0 & 1.0 \\
cnv         & 4.0 & 1.0 \\
\midrule
Total paid  & 19.0 & 6.0 \\
\bottomrule
\end{tabular}
\end{table}

\begin{table}[h]
\centering
\scriptsize
\caption{Cost schedules for the tabular medical datasets (Heart, Adult,
PhysioNet). Cube and MiniBooNE use unit cost 1.0 for all $K{-}1$ paid
modalities and 0.0 for the single free modality. Values in clinical or user-effort units.}
\label{tab:costs-tabular}
\begin{tabular}{lc @{\hspace{2em}} lc @{\hspace{2em}} lc}
\toprule
\multicolumn{2}{c}{\textbf{Heart}} &
\multicolumn{2}{c}{\textbf{Adult}} &
\multicolumn{2}{c}{\textbf{PhysioNet}} \\
Feature & Cost & Feature & Cost & Modality & Cost \\
\midrule
age        & 0.0 & age            & 0.0 & demographics  & 0.0 \\
sex        & 0.0 & sex            & 0.0 & vital\_basic  & 0.3 \\
cp         & 0.3 & race           & 0.2 & vital\_bp     & 0.3 \\
trestbps   & 0.5 & native-country & 0.3 & cbc           & 1.0 \\
chol       & 0.8 & marital        & 0.3 & glucose\_lytes & 1.0 \\
fbs        & 0.8 & relationship   & 0.3 & renal         & 1.0 \\
restecg    & 1.0 & education      & 0.8 & neuro\_vent   & 1.0 \\
thalach    & 2.0 & education-num  & 0.5 & lfts          & 3.0 \\
exang      & 2.0 & workclass      & 1.0 & abg           & 3.0 \\
oldpeak    & 2.0 & occupation     & 1.5 & cardiac       & 4.0 \\
slope      & 2.0 & hours-per-week & 1.5 & & \\
ca         & 3.5 & capital-gain   & 3.0 & & \\
thal       & 4.0 & capital-loss   & 3.5 & & \\
           &     & fnlwgt         & 0.5 & & \\
\bottomrule
\end{tabular}
\end{table}

\subsection{AFABench protocol: AFABench per-feature datasets}

\begin{table}[h]
\centering
\small
\caption{AFABench acquires one feature at a time, so
$K = D$ (number of features). Free tier: for Cube, the single free
feature $f_0$; for MiniBooNE, the single free feature $f_0$; for
PhysioNet, the shared classifier begins with no observed features (cold
start).}
\label{tab:datasets-b}
{\setlength{\tabcolsep}{4.5pt}%
\begin{tabular}{lcccl}
\toprule
Dataset & $D$ & $C$ & $N$ (per instance) & Notes \\
\midrule
Cube      & 12 & 8 & 1000 & Synthetic; $f_0$ free \\
MiniBooNE & 50 & 2 & 2000 & $f_0$ free \\
PhysioNet & 41 & 2 & 1108 & cold start (no free) \\
\bottomrule
\end{tabular}}
\end{table}

Per-feature cost is uniformly 1 in the AFABench protocol, so budget $B$
in Section~\ref{sec:experiments-source-b} equals the number of features that can be
acquired. Hard budgets evaluated are $B \in \{3, 5, 10\}$ for Cube and
$B \in \{5, 10, 15\}$ for MiniBooNE and PhysioNet.

\section{Additional Results}
\label{app:results}

\subsection{Training-recipe ablation}
\label{app:training-recipe}

The Method's training loss is
$\mathcal{L}_\text{act} + \lambda_\text{cls}\mathcal{L}_\text{cls}$
with (i) a soft-KL action loss over the full per-candidate Brier
ranking and (ii) an auxiliary CE regularizer on the shared trunk.
An earlier variant used categorical CE against the oracle's argmax
(no soft ranking) and no auxiliary CE. Under 5-fold CV, the current recipe beats the CE-on-argmax
variant by $+1.9$\,pp on average and is positive on every dataset
(Table~\ref{tab:training-recipe-a}). Distilling the full ranking extracts more of the teacher's signal than distilling only its argmax.  

\begin{table}[h]
\centering\small
\caption{Training-recipe ablation on the medical
modality-granularity setting (5-fold CV, mean-across-budgets
accuracy). Current recipe (soft-KL + auxiliary CE) improves over
the CE-on-argmax variant on all six datasets, with a 
mean improvement $+1.9$\,pp.}
\label{tab:training-recipe-a}
\begin{tabular}{lccc}
\toprule
Dataset & CE-argmax & soft-KL + aux CE & $\Delta$ \\
\midrule
adult      & 0.755 & 0.769 & $+0.014$ \\
cube       & 0.538 & 0.587 & $+0.049$ \\
heart      & 0.608 & 0.612 & $+0.004$ \\
glioma     & 0.798 & 0.810 & $+0.012$ \\
miniboone  & 0.849 & 0.854 & $+0.005$ \\
physionet  & 0.640 & 0.669 & $+0.029$ \\
\midrule
mean       & 0.698 & 0.717 & $+0.019$ \\
\bottomrule
\end{tabular}
\end{table}

\subsection{Example acquisition trajectories}
\label{app:trajectories}

Table~\ref{tab:trajectories} illustrates representative acquisition
trajectories on the Glioma cohort. Two qualitative patterns
emerge. First, patients share a common low-cost opening
(clinical $\rightarrow$ reports), reflecting universally informative early acquisitions. Second, trajectories become increasingly personalized as the budget grows, with different patients
selecting different modality sequences.
\begin{table*}[h]
\centering
\small
\caption{Acquisition trajectories for three TCGA-Glioma patients
(fold~0, seed~1) under the complexity-based cost schedule:
clinical free, reports $2.0$, histology $2.0$, mutations $3.0$,
miRNA / methylation / CNV $4.0$.}
\label{tab:trajectories}
\begin{tabular}{llll}
\toprule
Patient & Budget & Spent & Acquisition sequence \\
\midrule
\multirow{4}{*}{A (class 2)}
  & 2.0  & 2.0  & clinical $\to$ reports \\
  & 5.0  & 5.0  & clinical $\to$ reports $\to$ mutations \\
  & 9.0  & 9.0  & clinical $\to$ reports $\to$ CNV $\to$ mutations \\
  & 15.0 & 15.0 & clinical $\to$ reports $\to$ CNV $\to$ mutations $\to$ methylation $\to$ histology \\
\midrule
\multirow{4}{*}{B (class 2)}
  & 2.0  & 2.0  & clinical $\to$ reports \\
  & 5.0  & 4.0  & clinical $\to$ reports $\to$ histology \\
  & 9.0  & 8.0  & clinical $\to$ reports $\to$ CNV $\to$ histology \\
  & 15.0 & 15.0 & clinical $\to$ reports $\to$ CNV $\to$ histology $\to$ mutations $\to$ methylation \\
\midrule
\multirow{4}{*}{C (class 2)}
  & 2.0  & 2.0  & clinical $\to$ reports \\
  & 5.0  & 4.0  & clinical $\to$ reports $\to$ histology \\
  & 9.0  & 8.0  & clinical $\to$ reports $\to$ histology $\to$ CNV \\
  & 15.0 & 15.0 & clinical $\to$ reports $\to$ histology $\to$ CNV $\to$ mutations $\to$ methylation \\
\bottomrule
\end{tabular}
\end{table*}

\subsection{Full-range headline curves}
\label{app:fullrange}

\begin{figure*}[h]
\centering
\includegraphics[width=\linewidth]{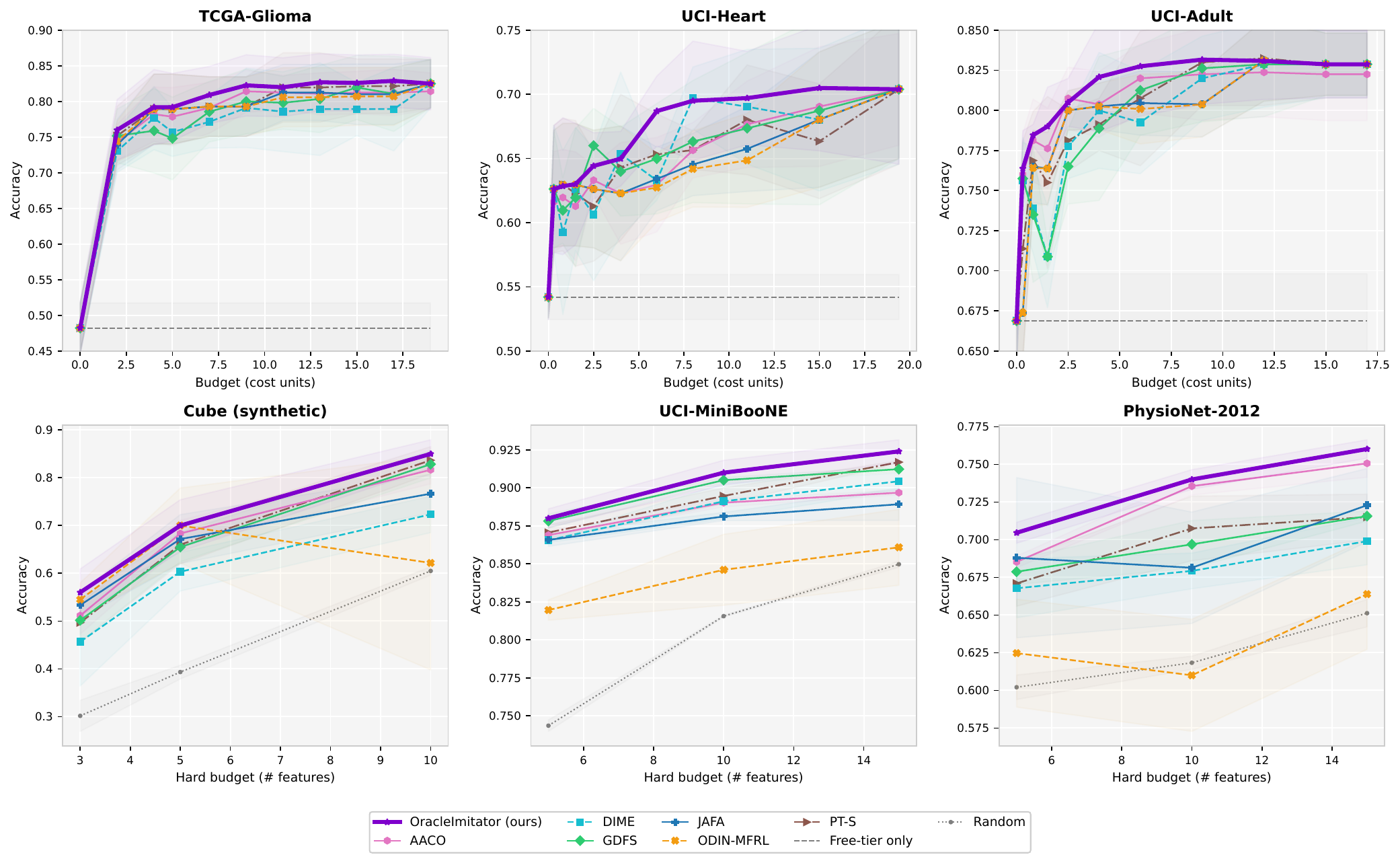}
\caption{Full-range companion to Fig.~\ref{fig:headline}: all six panels
at their full budget and accuracy range, with the free-tier floor and
random baseline visible. Same methods, predictors, protocols, styling, and
reported values as the main figure.}
\label{fig:headline-fullrange}
\end{figure*}

The main-body Figure~\ref{fig:headline} crops each panel to the
competitive accuracy band so that the differences between methods stay
legible. Figure~\ref{fig:headline-fullrange} shows the same six panels at their full budget and accuracy range.

\subsection{Cost robustness}

We evaluate robustness to the acquisition cost definition by varying
\emph{only} the cost schedule, holding the teacher, predictor, and training
protocol fixed.

\paragraph{Glioma: three cost schedules.}
\label{app:cost-robustness}

We compare the main complexity-based schedule (Appendix~\ref{app:datasets})
with two alternatives over the same acquisition units: a \emph{unit-cost}
schedule in which every paid unit costs $1.0$, and a fixed-seed
\emph{permutation} of the original costs that breaks the alignment between
cost and informativeness.

\begin{figure*}[h]
\centering
\includegraphics[width=\linewidth]{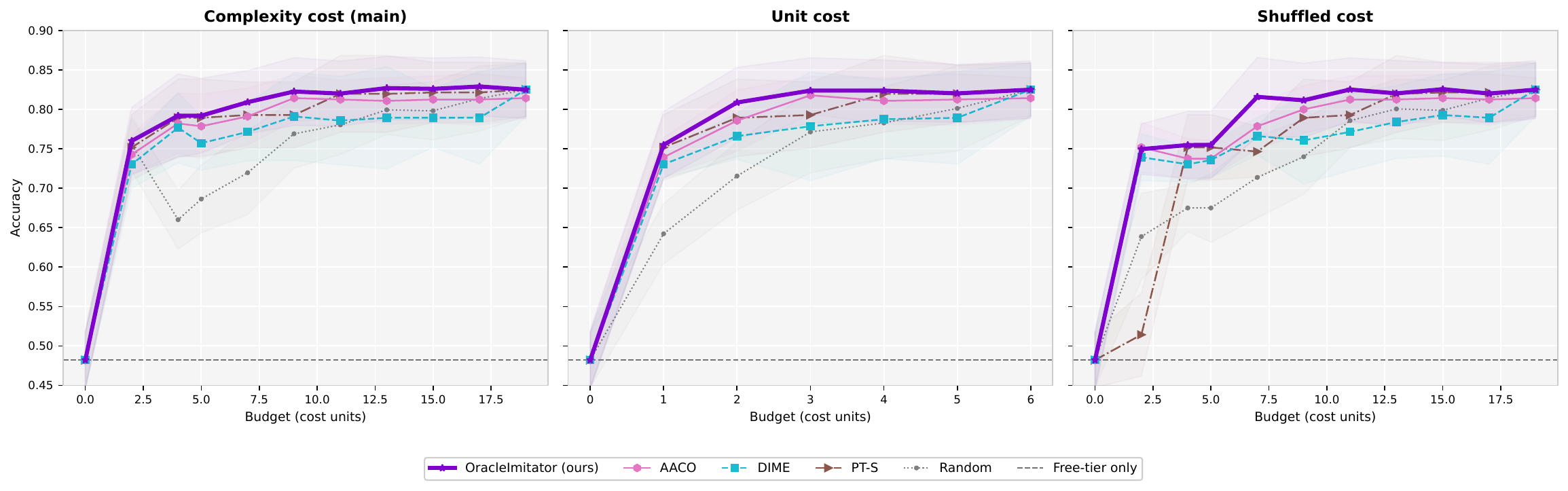}
\caption{\textbf{Glioma under three cost schedules.} Left: complexity-based
cost (main). Middle: unit cost. Right: fixed-seed shuffled cost. Same
methods, same 5-fold CV protocol, same late-fusion predictor.
All methods are aggregated identically as plain seed means (no
best-of-$n$ selection); shaded bands are $\pm 1$\,s.d.\ across the five
fold-level statistics. The acquisition trajectory changes with the schedule, but
\ourmethod\ remains the top method under all three.}
\label{fig:m7-cost-robustness}
\end{figure*}

\ourmethod\ ranks first under all three schedules, with mean-across-budgets
accuracy varying by less than $1$\,pp ($0.806$ / $0.809$ / $0.800$), while
the strongest baseline is substantially more sensitive to the permuted
costs ($0.802 \to 0.763$).

\paragraph{Tabular datasets: matched unit-cost variant.}
\label{app:cost-scheme}

We repeat the comparison on Heart, Adult, and PhysioNet under matched
unit-cost variants in which every non-free feature costs $1.0$
(\texttt{heart\_unit}, \texttt{adult\_unit}, \texttt{physionet\_unit}), with
all other settings held fixed. \ourmethod\ remains first on Heart and Adult
and is statistically tied on PhysioNet ($\Delta = -0.2$\,pp), preserving the
overall method ordering.

Overall, these results indicate that \ourmethod's advantage is robust to
both heterogeneous and uniform acquisition cost definitions.

\subsection{Predictor-Agnostic Ablation}
\label{app:predictor-agnostic}

We test whether \ourmethod\ depends on the late-fusion predictor by
replacing it with the shared mask-aware MLP used by AFABench, keeping the
acquisition pipeline unchanged, and rerunning the three medical setups
(Heart, Adult, Glioma). As shown in Table~\ref{tab:predictor-agnostic}, \ourmethod\ is robust across
substantially different downstream predictors.

\begin{table*}[h]
\centering\small
\caption{\textbf{Predictor-agnostic ablation.} Mean-across-budgets
accuracy per (dataset, method). LF = late-fusion stacking (used in main body), MM = shared mask-aware MLP. Parentheses give \ourmethod's rank among the four methods under that predictor.}
\label{tab:predictor-agnostic}
\begin{tabular}{lcccccccc}
\toprule
Dataset & \multicolumn{2}{c}{OracleImitator} & \multicolumn{2}{c}{DIME} & \multicolumn{2}{c}{AACO} & \multicolumn{2}{c}{Random} \\
        & LF & MM & LF & MM & LF & MM & LF & MM \\
\midrule
Heart   & \textbf{0.642 (1)} & \textbf{0.617 (1)} & 0.612 & 0.586 & 0.637 & 0.605 & 0.628 & 0.596 \\
Adult   & 0.791 (2)          & \textbf{0.758 (1)} & 0.760 & 0.746 & 0.792 & 0.743 & 0.752 & 0.737 \\
Glioma  & \textbf{0.804 (1)} & \textbf{0.797 (1)} & 0.781 & 0.782 & 0.797 & 0.781 & 0.761 & 0.744 \\
\midrule
Mean $\Delta$ (MM $-$ LF) & \multicolumn{2}{c}{$-0.022$} & \multicolumn{2}{c}{$-0.013$} & \multicolumn{2}{c}{$-0.032$} & \multicolumn{2}{c}{$-0.021$} \\
\bottomrule
\end{tabular}
\end{table*}

\section{Runtime and Scalability}
\label{app:complexity}

We characterize the training- and inference-time cost of \ourmethod\
under both medical setting (modality-granularity, late-fusion predictor) and
AFABench protocol (per-feature, AFABench shared masked-MLP predictor). The two
predictor architectures give qualitatively different scaling profiles
that are important for choosing between them.

\subsection{Downstream predictor implementations}
\label{app:predictors}

\paragraph{Medical setting.} We use a two-stage late-fusion predictor \citep{song2026multimodal}. 
Each modality is first mapped to class probabilities by a unimodal logistic
regressor; a subset-specific logistic regression then fuses the
probabilities of the available modalities. Fusion heads are fitted lazily
and cached by observed subset, so only the subsets actually visited are
ever fitted. In principle this yields up to $2^K$ distinct fusion heads. In
practice, the cache size is
$\sim\!  1$--$3$K subsets on our medical datasets ($K \le 12$),
well below the $2^{12} = 4096$ upper bound. Each fitted fusion logreg
is small ($|S| \cdot C \le 36$-dim input, $N \approx 450$ training
patients) and fits in $\sim\! 20$--$50$\,ms.

\paragraph{AFABench protocol.} We use the shared mask-aware MLP provided by
AFABench, $f_\theta(\bx_S, \bmm_S) = \text{MLP}([\bx \odot \bmm_S,\,
\bmm_S])$, where a single network handles all observed subsets through the
binary mask.

\subsection{Runtime}

Tables~\ref{tab:complexity-baselines} and ~\ref{tab:measured-runtime}: offline teacher generation takes
$138$--$405$\,s per fold and student training a further $7$--$16$\,s, with a
fusion-cache footprint under $7$\,MB. At deployment every learned policy costs
a few milliseconds per patient, dominated by the shared downstream predictor:
\ourmethod\ is comparable to DIME/GDFS at inference but substantially cheaper
to train than the RL baselines ($\sim\!10$ minutes versus $1$--$3$ hours).
AACO, though training-free, is the most expensive at deployment
($\sim\!66$\,ms) because it recomputes a per-patient score at each step.

\begin{table}[h]
\centering\small
\caption{\emph{Measured} deploy-time acquisition cost per patient on
Glioma (fold~0, single CPU core; mean over budgets with the
first warm-up step excluded), plus offline training cost. Acquisition
time is the policy-specific per-patient trajectory only; the final
downstream prediction adds a further $\sim\!7$--$17$\,ms shared by
every method (Table~\ref{tab:measured-runtime}), so total deploy time
is prediction-dominated and comparable across the learned policies.}
\label{tab:complexity-baselines}
\fitcol{\begin{tabular}{lcc}
\toprule
Method & Acquisition (ms/patient) & Offline training \\
\midrule
\ourmethod\ (ours)   & 5.5   & 5--10 min \\
DIME                  & 5.3   & 30--60 min \\
GDFS                  & 5.3   & 30--60 min \\
JAFA (RL)             & 2.7   & 1--3 h \\
ODIN (RL)             & 9.5   & 1--3 h \\
AACO                  & 65.9  & 0 \\
Permutation (PT-S)    & 0.02  & 0 \\
\bottomrule
\end{tabular}}
\end{table}

\begin{table*}[h]
\centering\small
\caption{Measured runtime and cache memory per medical dataset
(fold~0). Times in seconds unless noted. \emph{T\_teach report} and
\emph{T\_student report} are extrapolated to 100 epochs from a
5-epoch steady-state measurement.}
\label{tab:measured-runtime}
\fittext{\begin{tabular}{lccccccccc}
\toprule
Dataset & $N_\text{train}$ & $K$ & T\_pred & T\_teach/ep & T\_teach 100ep & T\_student 100ep & T\_infer/patient & Cache subsets & Cache size \\
\midrule
Heart          & 237  & 13 & 0.04 s & 2.17 s & 217 s & 6.9 s & 17.1 ms & 1812 & 991 KB \\
Adult          & 640  & 14 & 0.03 s & 3.65 s & 365 s & 11.2 s & 17.3 ms & 3845 & 509 KB \\
Cube           & 800  & 12 & 0.05 s & 3.94 s & 394 s & 12.6 s & 15.3 ms & 2046 & 6.6 MB \\
MiniBooNE      & 1200 & 12 & 0.03 s & 4.05 s & 405 s & 15.8 s & 14.0 ms & 2048 & 224 KB \\
PhysioNet      & 886  & 10 & 0.02 s & 2.27 s & 227 s & 11.7 s & 12.5 ms & 512  & 48 KB \\
Glioma (M=7)                & 448  & 7  & 0.34 s & 1.38 s & 138 s & 10.2 s & 17.4 ms & 64   & 24 KB \\
Glioma (12-mod.\ stress test) & 448  & 12 & 0.19 s & 2.28 s & 228 s & 8.6 s  & 14.2 ms & 1993 & 956 KB \\
\bottomrule
\end{tabular}}
\end{table*}

\subsection{Scalability}
\label{app:complexity-scaling}

\begin{figure*}[h]
\centering
\includegraphics[width=0.85\linewidth]{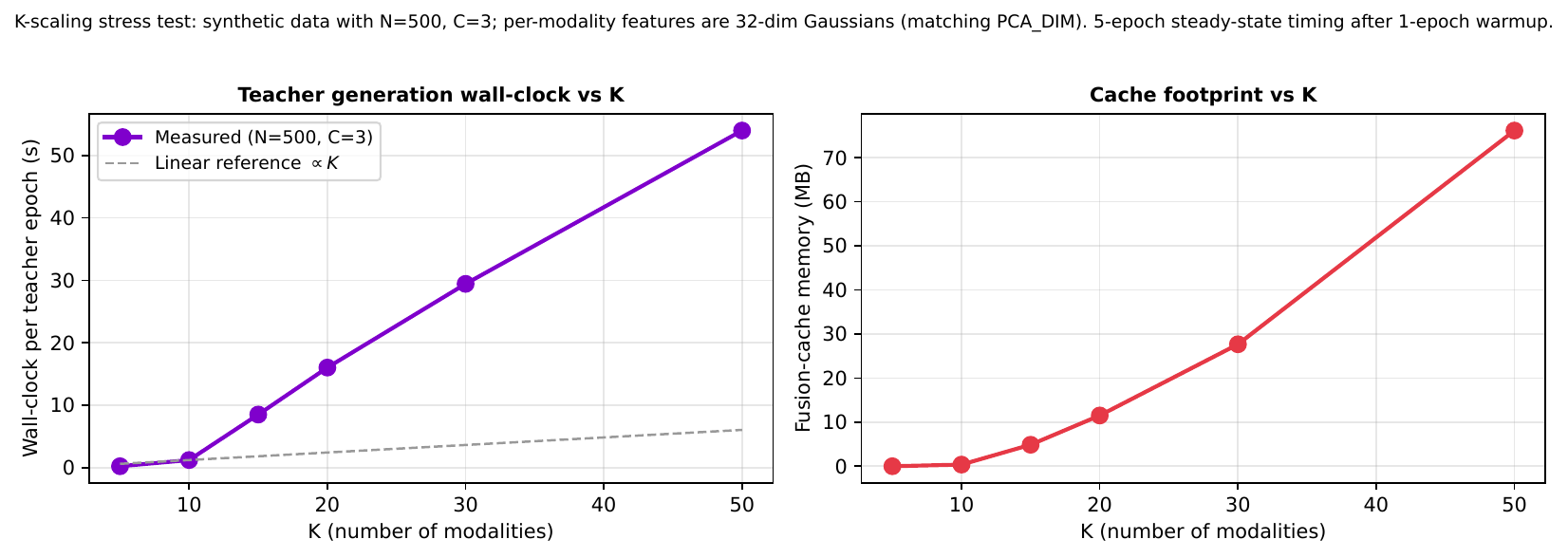}
\caption{\textbf{K-scaling stress test.} Left: teacher-generation
wall-clock per epoch as $K$ grows. The dashed grey line is a linear
reference $\propto K$ anchored at $K = 10$; the measured curve tracks
approximately linear through $K \le 20$ and grows super-linearly beyond
because per-fusion-head fits become larger with $|S|$ and the fusion
cache stops catching most of the encountered subsets. Right: cache
footprint (MB) as $K$ grows. Even at $K = 50$ the cache stays under
$80$\,MB, which is orders of magnitude smaller than the memory of any
learned baseline (DIME/GDFS masked classifiers, EDDI's partial VAE)
that would occupy hundreds of MB or more.}
\label{fig:k-scaling}
\end{figure*}

Figure~\ref{fig:k-scaling} shows that teacher generation scales
approximately linearly with $K$ up to $K \approx 20$, after which
per-subset late-fusion fitting becomes the wall-clock bottleneck. This is
why the per-feature AFABench experiments (effective $K = D = 50$ for
MiniBooNE) use the shared mask-aware predictor, which handles all $2^K$
subsets with one parameter set. In the modality-granularity regime we
target ($K \le 14$), late fusion remains firmly in the linear regime.

\label{app:teacher-pruning}%
For larger candidate sets, retrieval-based candidate pruning further
reduces teacher computation: we optionally pre-select the top $M$
candidates with a cheap retrieval heuristic before computing the exact
Brier scores, so the expensive fusion-head lookup is triggered only $M$
times per sample (Algorithm~\ref{alg:teacher-pruning}). Pruning to
$M = 5$ costs at most $0.7$\,pp of accuracy ($M = 3$ costs up to
$1.4$\,pp), indicating that the teacher signal the student needs is
contained in the top few candidates rather than the full scan. We do not
use pruning in the main experiments ($K \le 14$ makes it unnecessary).

\begin{algorithm}[h]
\caption{Retrieval-Based Candidate Pruning for Teacher Acceleration}
\label{alg:teacher-pruning}
\begin{algorithmic}[1]
\REQUIRE Sample $(x_i, y_i)$, observed subset $S$, prune size $M$,
neighborhood size $k$, predictor $f_\theta$, training set
$\{(x_j, y_j)\}$.
\STATE \COMMENT{Phase 1: cheap retrieval-based ranking (no fusion-head calls)}
\STATE $X_\text{pool} \gets \text{concat}(\{x_j^S\}_j)$
        \COMMENT{cached from stage-1 unimodal features}
\STATE $x_q \gets x_i^S$
\STATE $d_j \gets \|x_j^S - x_q\|_2 \quad \forall j$
\STATE $d_i \gets +\infty$
        \COMMENT{leave-one-out}
\STATE $\mathcal N \gets \text{top-}k \text{ indices of } d$
\STATE $\text{rank}[a] \gets \sum_d \text{Var}_{j \in \mathcal N}(x_j^{(a)})[d]$
        for every candidate $a \notin S$
        \COMMENT{high variance across neighbors $\Rightarrow$ candidate is heterogeneous
        $\Rightarrow$ potentially informative}
\STATE $\text{kept} \gets \text{top-}M \text{ candidates by } \text{rank}$
\STATE \COMMENT{Phase 2: exact Brier only on the kept candidates}
\STATE $\text{score}[a] \gets +\infty$ for every candidate $a$
\STATE $y_\text{oh} \gets \text{one\_hot}(y_i)$
\FOR{each $a \notin S$}
    \IF{$a \notin \text{kept}$}
        \STATE \textbf{continue}
        \COMMENT{pruned; leaves score at $+\infty$ so its softmax mass is 0}
    \ENDIF
    \STATE $p_a \gets f_\theta(x_i^{S \cup \{a\}})$
        \COMMENT{expensive: fusion-head evaluation}
    \STATE $\text{score}[a] \gets \|p_a - y_\text{oh}\|^2$
        \COMMENT{exact Brier}
\ENDFOR
\STATE $q_S(a) \gets \exp(-\text{score}[a] / T) / \sum_{a'} \exp(-\text{score}[a'] / T)$
        \COMMENT{softmax teacher target; pruned candidates get 0 mass}
\RETURN $q_S$
\end{algorithmic}
\end{algorithm}

\begin{table}[h]
\centering\small
\caption{Candidate-pruning teacher acceleration. Student mean-across
-budgets accuracy under 5-fold CV with 3 seeds per fold. Pruning to
$M = 5$ costs $\le 0.7$\,pp of accuracy while replacing $K = 12$
per-candidate teacher forward passes with $M = 5$; pruning to $M = 3$
costs $\le 1.4$\,pp.}
\label{tab:teacher-pruning}
\fitcol{\begin{tabular}{lcccc}
\toprule
Dataset & $M = 3$ & $M = 5$ & $M = $ all & $\Delta_{M=5}$ vs.\ all \\
\midrule
Cube      & 0.590 $\pm$ 0.023 & 0.590 $\pm$ 0.019 & 0.594 $\pm$ 0.016 & $-0.004$ \\
Glioma-12 (unit)  & 0.778 $\pm$ 0.032 & 0.785 $\pm$ 0.030 & 0.784 $\pm$ 0.031 & $+0.001$ \\
Glioma-12 (info)  & 0.702 $\pm$ 0.028 & 0.709 $\pm$ 0.025 & 0.716 $\pm$ 0.028 & $-0.007$ \\
\bottomrule
\end{tabular}}
\end{table}
\section{Mechanism Analysis}
\label{app:mechanism}

\subsection{Recoverability diagnostics}
\label{app:extra}

\begin{table}[h]
\centering\small
\caption{$k$NN teacher-learnability diagnostic. Higher consistency indicates that more of the teacher's
action is recoverable from deployment-visible observations; the
\emph{Random} column gives the chance agreement $1/K_\text{paid}$ for each
action space.}
\label{tab:identifiability}
\begin{tabular}{lcccc}
\toprule
Dataset  & $K_\text{paid}$ & Consistency $\uparrow$ & Norm.\ $H \downarrow$ & Random \\
\midrule
cube     & 11 & 0.297 & 0.699 & 0.091 \\
adult    & 12 & 0.385 & 0.571 & 0.083 \\
heart    & 11 & 0.418 & 0.563 & 0.091 \\
miniboone& 11 & 0.428 & 0.574 & 0.091 \\
physionet&  9 & 0.444 & 0.596 & 0.111 \\
glioma   &  6 & 0.513 & 0.612 & 0.167 \\
\bottomrule
\end{tabular}
\end{table}

Table~\ref{tab:identifiability} reports the diagnostic at the bottleneck
free-tier state, where every dataset sits far above its chance baseline but
well below $1$.

\subsection{Continuity support for the identifiability estimator}
\label{sec:continuity-support}

Formally, the estimator is justified when the mapping
$\bx_S \mapsto P(y \mid \bx_S)$ is locally continuous in the $\bx_S$-embedding
used for the $k$NN lookup, so that a $k$NN estimate approximates what any
Lipschitz-continuous predictor achieves locally; C1 and C2 probe the two
ways this local continuity can fail.

\paragraph{(C1) Do close-embedding pairs share labels more often?}
For each of 20\,K sampled patient pairs $(i, j)$ we compute the
Euclidean distance $\|\bx_i^S - \bx_j^S\|$ and the label-agreement
indicator $\mathbf{1}[y_i = y_j]$, then take the Pearson correlation.
Under label-respecting embedding, closer pairs should be more
likely to share a label (negative Pearson).

\paragraph{(C2) Does the predictor track the local $k$NN
distribution?} For each training patient $i$ we compute the soft
$k$-NN label distribution $\widehat P_{k\text{NN}}(y \mid \bx_i^S)$
and our late-fusion predictor's output
$\widehat P_\theta(y \mid \bx_i^S)$, then correlate the two.
Under Lipschitz continuity, any consistent predictor should
approximate a local kernel estimate; a high correlation is
empirical evidence for that.

\begin{table*}[h]
\centering\small
\caption{Continuity diagnostics on 6 datasets $\times$ 3
observed-subset states, mean~$\pm$ s.d.\ across 5 CV folds. C1:
pairwise Pearson $r$(embedding-distance, same-label-indicator) on
20\,K sampled pairs (expected sign: negative). C2 (flat): Pearson
$r$ between $\widehat P_{k\text{NN}}(y \mid \bx_S)$ and
$\widehat P_\theta(y \mid \bx_S)$, flattened over patients $\times$
classes ($k = 10$). C2 (argmax): fraction of patients where the
two distributions agree on argmax.}
\label{tab:continuity}
\begin{tabular}{llccc}
\toprule
Dataset & State & C1: $r$(dist, same) & C2 flat: $r$($P_{k\text{NN}}$, $P_\theta$) & C2 argmax \\
\midrule
Heart          & free          & $+0.006 \pm 0.005$ & $+0.792 \pm 0.014$ & $0.761 \pm 0.032$ \\
Heart          & mid           & $-0.101 \pm 0.026$ & $+0.882 \pm 0.007$ & $0.775 \pm 0.029$ \\
Heart          & near-terminal & $-0.229 \pm 0.011$ & $+0.922 \pm 0.008$ & $0.824 \pm 0.007$ \\
\midrule
Adult          & free          & $-0.081 \pm 0.010$ & $+0.672 \pm 0.012$ & $0.815 \pm 0.032$ \\
Adult          & mid           & $-0.073 \pm 0.007$ & $+0.742 \pm 0.015$ & $0.811 \pm 0.017$ \\
Adult          & near-terminal & $-0.112 \pm 0.011$ & $+0.887 \pm 0.008$ & $0.863 \pm 0.011$ \\
\midrule
Cube           & free          & $-0.003 \pm 0.008$ & $+0.104 \pm 0.018$ & $0.148 \pm 0.016$ \\
Cube           & mid           & $-0.143 \pm 0.004$ & $+0.802 \pm 0.005$ & $0.382 \pm 0.019$ \\
Cube           & near-terminal & $-0.349 \pm 0.009$ & $+0.937 \pm 0.003$ & $0.881 \pm 0.012$ \\
\midrule
MiniBooNE      & free          & $-0.298 \pm 0.006$ & $+0.935 \pm 0.005$ & $0.909 \pm 0.007$ \\
MiniBooNE      & mid           & $-0.219 \pm 0.015$ & $+0.933 \pm 0.004$ & $0.918 \pm 0.007$ \\
MiniBooNE      & near-terminal & $-0.191 \pm 0.009$ & $+0.911 \pm 0.002$ & $0.905 \pm 0.006$ \\
\midrule
PhysioNet      & free          & $-0.022 \pm 0.008$ & $+0.537 \pm 0.061$ & $0.682 \pm 0.038$ \\
PhysioNet      & mid           & $-0.011 \pm 0.009$ & $+0.543 \pm 0.066$ & $0.706 \pm 0.011$ \\
PhysioNet      & near-terminal & $-0.030 \pm 0.004$ & $+0.613 \pm 0.008$ & $0.701 \pm 0.005$ \\
\midrule
Glioma         & free          & $-0.061 \pm 0.010$ & $+0.806 \pm 0.019$ & $0.650 \pm 0.040$ \\
Glioma         & mid           & $-0.096 \pm 0.008$ & $+0.771 \pm 0.005$ & $0.727 \pm 0.018$ \\
Glioma         & near-terminal & $-0.073 \pm 0.009$ & $+0.625 \pm 0.014$ & $0.637 \pm 0.013$ \\
\bottomrule
\end{tabular}
\end{table*}

\paragraph{Table~\ref{tab:continuity}.} C1 is negative in nearly all states and becomes stronger as more modalities are observed, indicating increasing local label consistency. C2 is high for most datasets (typically $r \ge 0.8$ at mid-to-late acquisition states), showing strong agreement between the predictor and the local $k$NN estimate. The only notable exception is the near-terminal glioma state, where the correlation drops, consistent with the known degradation of Euclidean $k$NN in high dimensions.

\subsection{Architecture ablation}
\label{app:architecture}

We ablate the imitator trunk across four architectures. We compare the main MLP against a wider MLP, a transformer,
a cross-attention model(Table~16). Each dataset is evaluated at a single fixed
budget: the evaluated budget nearest the midpoint of that dataset's budget range (column $B$ in Table~\ref{tab:arch-sweep}). 

\begin{table*}[h]
\centering\small
\caption{Total gap between the oracle and the CV imitator, and the
share of that gap that the in-sample imitator closes. If the closed
share is high, more data / regularization can approach the oracle; if
it is low, the residual gap is architectural / predictor-side.}
\label{tab:insample}
\begin{tabular}{lccccc}
\toprule
Dataset  & Oracle & imit (CV) & imit (in-sample) & gap\_total & gap\_closed \\
\midrule
Cube      & 0.990 & 0.678 & 0.774 & 0.312 & $+0.096$ (31\%) \\
Heart     & 0.849 & 0.636 & 0.701 & 0.212 & $+0.064$ (30\%) \\
PhysioNet & 0.934 & 0.723 & 0.779 & 0.211 & $+0.056$ (27\%) \\
Glioma    & 0.954 & 0.823 & 0.893 & 0.131 & $+0.070$ (54\%) \\
Adult     & 0.970 & 0.809 & 0.840 & 0.161 & $+0.031$ (19\%) \\
MiniBooNE & 0.956 & 0.857 & 0.873 & 0.099 & $+0.015$ (16\%) \\
\bottomrule
\end{tabular}
\end{table*}

\begin{table*}[h]
\centering\small
\caption{Architecture sweep at a fixed mid-budget point. Alternative student architectures produce only marginal differences from the main \texttt{imit} model, while the in-sample upper bound (\texttt{in-samp}) remains substantially higher across all datasets.}
\label{tab:arch-sweep}
\begin{tabular}{lccccccc}
\toprule
Dataset  & $B$ & Oracle & \texttt{imit} & \texttt{in-samp} & \texttt{mlp\_big} & \texttt{xfmr} & \texttt{xattn} \\
\midrule
Cube      & 4 & 0.990 & 0.678 & \textbf{0.774} & 0.711 & 0.703 & 0.685 \\
Heart     & 4 & 0.849 & 0.636 & \textbf{0.701} & 0.643 & 0.633 & 0.636 \\
PhysioNet & 4 & 0.934 & 0.723 & \textbf{0.779} & 0.728 & 0.725 & 0.741 \\
Glioma    & 9 & 0.954 & 0.823 & \textbf{0.893} & 0.829 & 0.815 & 0.838 \\
Adult     & 4 & 0.970 & 0.809 & \textbf{0.840} & 0.821 & 0.821 & 0.828 \\
MiniBooNE & 4 & 0.956 & 0.857 & \textbf{0.873} & 0.852 & 0.853 & 0.847 \\
\bottomrule
\end{tabular}
\end{table*}

No architecture significantly beats the main \texttt{imit}. The best
average delta ($+0.5$\,pp for \texttt{xattn}) is within seed noise. Simultaneously, the in-sample column
dominates every one of these architectures by 4--6\,pp on average.
This is decisive: if capacity or inductive bias were the bottleneck,
at least one of \texttt{mlp\_big}, \texttt{xfmr}, or \texttt{xattn}
would close some of the gap to the in-sample column. None do.
\subsection{Why the medical-setting margin exceeds the AFABench margin}
\label{app:why-late-fusion}

The separation between methods is wider under the late-fusion predictor
than under AFABench's shared mask-aware MLP. We attribute this to three
factors, of which the last two are specific to late fusion.

\paragraph{(1) Teacher--predictor alignment.} The teacher's Brier signal is
computed through the exact predictor $f_\theta$ used at evaluation, whereas
the discriminative baselines (DIME, GDFS) learn a value head tied to their
own masked classifier. This holds in both settings and is therefore a
precondition rather than the differentiator.

\paragraph{(2) Late fusion amplifies the value of a correct subset.} Each
observed subset is served by a separately fitted fusion head with no
parameter sharing, so a suboptimal acquisition cannot be smoothed over by a
shared representation; AFABench's mask-aware MLP shares one parameter set
across all $2^K$ subsets and compresses good and bad policies toward one
another.

\paragraph{(3) Late fusion is order-insensitive.} The fusion head depends
only on \emph{which} modalities are observed, not the order in which they
were acquired, reducing AFA to pure subset selection --- structurally the
problem our greedy Brier teacher solves. RL baselines model sequential
dynamics that are wasted in this regime; empirically JAFA and ODIN converge
to the same acquired subset, differing only in order.

\end{document}